# Modélisation Dynamique d'un Robot Parallèle à 3-DDL : l'Orthoglide

Sylvain GUEGAN, Wisama KHALIL, Damien CHABLAT, Philippe WENGER

Institut de Recherche en Communications et Cybernétique de Nantes
BP 92101  1, rue de la Noë, 44321 Nantes Cedex 03, France

wisama.khalil@irccyn.ec-nantes.fr
http://www.irccyn.ec-nantes.fr

*Résumé* — Dans cet article, nous proposons une méthode pour le calcul des modèles dynamiques inverse et direct de l'Orthoglide, un robot parallèle à trois degrés de liberté en translation. Ces modèles sont calculés à partir des éléments du modèle dynamique de la structure d'une chaîne cinématique et des équations de Newton-Euler appliquées à la plate-forme. Ces modèles sont obtenus sous forme explicite ayant une interprétation physique intéressante.

*Mots clés* — Robots parallèles, 3-DDL, dynamique, structures fermées, modèle dynamique inverse, modèle dynamique direct.

## I. Introduction

Les robots parallèles sont des systèmes multi-corps complexes, qui sont parmi ceux les plus difficiles à modéliser, à cause de leur architecture parallèle qui comporte plusieurs boucles fermées. Suite à l'augmentation constante des performances attendues par ce type de machines, la conception de leur commande doit prendre en compte les forces d'interactions dynamiques. D'où l'intérêt d'avoir un modèle dynamique efficace pour la commande en ligne.

Afin d'obtenir le modèle dynamique des robots parallèles, beaucoup de méthodes calculent le modèle dynamique de la structure arborescente et utilisent les multiplicateurs de Lagrange afin d'obtenir le modèle dynamique complet du robot [1-5]. Le principe des travaux virtuels a été utilisé dans [6,7]. La formulation de Newton-Euler a aussi été utilisée, par exemple : Reboulet et al. [8] ont donné une forme matricielle pour les robots parallèles de type Stewart, cependant leur modèle n'est pas complet. Ils négligent notamment la masse des pistons et la rotation autour de l'axe principal de chaque chaîne. Gosselin [9] a proposé le modèle dynamique inverse du robot Stewart dans lequel toutes les masses et inerties sont prises en compte, le problème direct n'a pas été traité. Dasgupta et al. [10] ont appliqué cette méthode à plusieurs robots parallèles de type planaires et spatiaux [11]. Ji [12] a étudié l'influence de l'inertie des chaînes cinématiques dans le modèle dynamique.

Cet article propose une solution pour la formulation des modèles dynamiques complets inverse et direct des robots parallèles. Ces modèles sont obtenus en termes des éléments du modèle dynamique cartésien des chaînes cinématiques du robot perçues aux points de connexions des chaînes avec la plate-forme. Par conséquent, on peut appliquer les techniques développées pour les robots séries aux calculs de ces modèles.

Nous avons récemment proposé une nouvelle méthode pour la modélisation dynamique du robot à 6 degrés de liberté Gough-Stewart [13]. Dans cet article, nous considérons l'application de cette méthode au robot Orthoglide [14].

L'architecture de l'Orthoglide est donnée sur la figure 1. La description des chaînes cinématiques est présentée sur la figure 2. L'Orthoglide est une machine de type parallèle possédant 3 articulations prismatiques orthogonales. La plate-forme mobile est connectée aux articulations prismatiques par 3 parallélogrammes articulés et bouge dans l'espace cartésien x-y-z avec une orientation fixe.

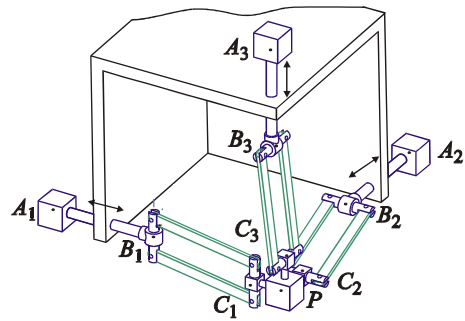

Fig. 1. : Architecture de l'Orthoglide

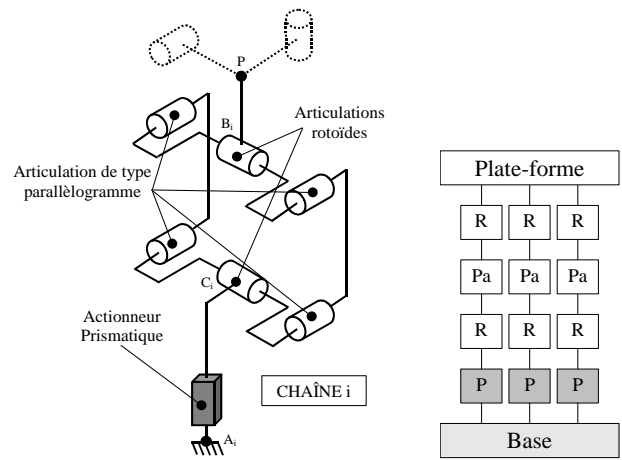

Fig. 2. : Description du robot Orthoglide

L'Orthoglide est dédié à l'usinage à grande vitesse, car son architecture se rapproche des machines standards d'architecture série PPP (espace de travail Cartésien régulier et performances uniformes) et avec, en plus, les propriétés des structures parallèles (inerties moins importantes et meilleures performances dynamiques). Son espace de travail est proche d'un cube et ne possède aucune singularité. Il existe une configuration où la matrice Jacobienne est isotrope avec tous ces facteurs d'amplification de vitesse égaux à 1. Ces facteurs restent bornés dans l'intervalle [1/2 ; 2] dans le reste de l'espace de travail.

Le paragraphe suivant présente la description géométrique du robot. Le paragraphe trois traite de la modélisation cinématique du robot. Les paragraphes quatre et cinq donnent respectivement les modèles dynamiques inverse et direct du robot. Un paragraphe final permet de tirer les conclusions sur ce travail.

## II. DESCRIPTION DU ROBOT

L'Orthoglide est un robot parallèle à trois degrés de liberté en translation. Il est composé d'une plate-forme mobile et de trois chaînes cinématiques identiques. Chaque chaîne est composée d'un actionneur prismatique (P) liant la base à la chaîne (point $A_i$ pour i = 1, 2, 3) d'une articulation rotoïde (R), d'une articulation de type parallélogramme (Pa) et d'une articulation rotoïde liant la chaîne à la plate-forme (figure 2).

Le robot a une structure complexe avec 2 boucles spatiales et 3 boucles planaires. La structure arborescente équivalente est obtenue en isolant la plate-forme et en coupant les trois articulations passives $q_{8i}$ (i = 1, 2, 3) (figure 3) :

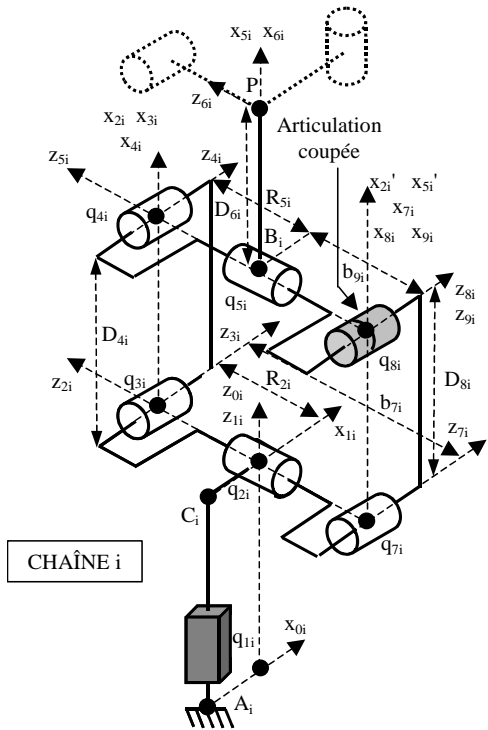

Fig. 3. : Placement des repères sur la chaîne i

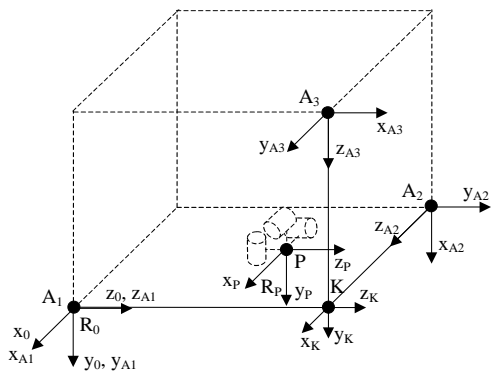

Fig. 4. : Repère de la base $R_0$ et repère de la plate-forme $R_P$

Les axes des trois actionneurs se coupent au point K. Le repère fixe $R_K$ a pour origine le point K. Les axes $x_K$, $y_K$, $z_K$ sont définis par les axes des trois actionneurs ($q_{1i}$), respectivement $q_{12}$, $q_{13}$, $q_{11}$. Nous définissons également deux repères : le repère $R_0$ fixe par rapport à la base et le repère $R_p$ fixe par rapport à la plate-forme, respectivement d'origine $A_1$ et P. Leurs axes, respectivement ($x_0$, $y_0$, $z_0$) et ($x_P$, $y_P$, $z_P$), sont parallèles aux axes ($x_k$, $y_k$, $z_k$). De plus, nous introduisons 3 repères $R_{Ai}$ d'origine $A_i$. Leurs axes sont définis de la manière suivante : l'axe $z_{Ai}$ est le long de l'axe de l'actionneur i et le plan ($x_{Ai}$, $z_{Ai}$) est défini pour i = 1 à 3 respectivement par : ($A_1$, K, $A_2$), ($A_2$, K, $A_3$), ($A_3$, K, $A_2$). La figure 4 indique l'emplacement de ces repères.

La notation de Khalil et Kleinfinger [15] peut-être utilisée pour décrire la géométrie de la structure arborescente. Le placement des repères est indiqué sur la figure 3. On peut remarquer que l'on utilise deux repères $R_{8i}$ et $R_{9i}$, pour l'articulation coupée $q_{8i}$ [15]. Les paramètres géométriques nécessaires pour définir le repère du 1$^{er}$ corps de chaque chaîne $R_{1i}$, dans le repère de base du robot $R_0$, sont donnés dans le tableau 1. Le tableau 2 donne la description du reste de chaque chaîne :

| $j_i$ | $a(j_i)$ | $\mu_{ji}$ | $\sigma_{ji}$ | $\gamma_{ji}$ | $b_{ji}$ | $\alpha_{ji}$ | $d_{ji}$ | $\theta_{ji}$ | $r_{ji}$ |
|---|---|---|---|---|---|---|---|---|---|
| $1_1$ | 0 | 1 | 1 | 0 | 0 | 0 | 0 | 0 | $q_{11}$ |
| $1_2$ | 0 | 1 | 1 | $\pi/2$ | a | $\pi/2$ | 0 | 0 | $-a+q_{12}$ |
| $1_3$ | 0 | 1 | 1 | 0 | a | $-\pi/2$ | 0 | $-\pi/2$ | $-a+q_{13}$ |

Tableau 1 : Paramètres géométriques du repère du 1$^{er}$ corps de la chaîne i (pour i = 1 à 3)

| $j_i$ | $a(j_i)$ | $\mu_{ji}$ | $\sigma_{ji}$ | $\gamma_{ji}$ | $b_{ji}$ | $\alpha_{ji}$ | $d_{ji}$ | $\theta_{ji}$ | $r_{ji}$ |
|---|---|---|---|---|---|---|---|---|---|
| $2_i$ | $1_i$ | 0 | 0 | 0 | 0 | $-\pi/2$ | 0 | $q_{2i}$ | $r_{2i}$ |
| $3_i$ | $2_i$ | 0 | 0 | 0 | 0 | $-\pi/2$ | 0 | $q_{3i}$ | 0 |
| $4_i$ | $3_i$ | 0 | 0 | 0 | 0 | 0 | $D_{4i}$ | $q_{4i}$ | 0 |
| $5_i$ | $4_i$ | 0 | 0 | 0 | 0 | $\pi/2$ | 0 | $q_{5i}$ | $r_{5i}$ |
| $6_i$ | $5_i$ | 0 | 0 | 0 | 0 | 0 | $D_{6i}$ | 0 | 0 |
| $7_i$ | $2_i$ | 0 | 0 | 0 | $b_{7i}$ | $-\pi/2$ | 0 | $q_{7i}$ | 0 |
| $8_i$ | $7_i$ | 0 | 0 | 0 | 0 | 0 | $D_{8i}$ | $q_{8i}$ | 0 |
| $9_i$ | $5_i$ | 0 | 0 | 0 | $b_{9i}$ | $-\pi/2$ | 0 | 0 | 0 |

Remarque : $b_{7i} = -2 r_{2i}$, $b_{9i} = -r_{2i}$, $r_{5i} = -r_{2i}$, $D_{8i} = D_{4i}$
$q_{7i} = q_{3i}$, $q_{8i} = q_{4i} = -q_{3i}$, $q_{5i} = -q_{2i} - \pi/2$

Tableau 2 : Paramètres géométriques de la chaîne i

Les Paramètres des tableaux 1 et 2 sont :
a(j) qui donne le repère antécédent au repère j,
$\mu$(j) et $\sigma$(j) qui décrivent le type de l'articulation :
- $\mu$(j) = 1 si l'articulation j est motorisée et $\mu$(j) = 0 si elle est passive,
- $\sigma$(j) = 1 si l'articulation est prismatique et $\sigma$(j) = 0 si elle est rotoïde.

Les paramètres ($\gamma_j$, $b_j$, $\alpha_j$, $d_j$, $\theta_j$, $r_j$) sont utilisés pour définir le repère $R_j$ dans le repère de son antécédent $R_i$. La matrice de transformation composée de ces paramètres est notée :

$$^i T_j = \begin{bmatrix} ^i A_j & ^i P_j \\ 0_{(1 \times 3)} & 1 \end{bmatrix} \quad (1)$$

où :
$^i A_j$    est la matrice (3×3) qui définit l'orientation du repère $R_j$ dans le repère $R_i$.
$^i P_j$    est la matrice de position (3×1) qui définit l'origine du repère $R_j$ dans le repère $R_i$.

## III. MODÉLISATION CINÉMATIQUE

Les notations suivantes sont utilisées :

- $(X_p, Y_p, Z_p)$ coordonnées du point P exprimées dans le repère $R_0$
- $(X_{Ai}, Y_{Ai}, Z_{Ai})$ coordonnées du point $A_i$ exprimées dans le repère $R_0$
- **L** matrice composée des variables prismatiques du robot
  $\mathbf{L} = \begin{bmatrix} q_{11} & q_{12} & q_{13} \end{bmatrix}^T$
- $\dot{\mathbf{L}}$ matrice composée des vitesses des articulations prismatiques du robot : $\dot{\mathbf{L}} = \begin{bmatrix} \dot{q}_{11} & \dot{q}_{12} & \dot{q}_{13} \end{bmatrix}^T$
- $\dot{\mathbf{q}}_i$ matrice composée des vitesses des articulations indépendantes de la chaîne i : $\dot{\mathbf{q}}_i = \begin{bmatrix} \dot{q}_{1i} & \dot{q}_{2i} & \dot{q}_{3i} \end{bmatrix}^T$
- $\ddot{\mathbf{q}}_i$ matrice composée des accélérations des articulations indépendantes de la chaîne i : $\ddot{\mathbf{q}}_i = \begin{bmatrix} \ddot{q}_{1i} & \ddot{q}_{2i} & \ddot{q}_{3i} \end{bmatrix}^T$
- $^0\mathbf{V}_p$ vitesse linéaire de l'origine de la plate-forme,
- $^0\dot{\mathbf{V}}_p$ accélération linéaire de l'origine de la plate-forme.

Les modèles suivants sont présentés dans [16,17] et ont une solution unique :

i) Le modèle géométrique inverse du robot donne les variables prismatiques motorisées ($q_{1i}$) en fonction des coordonnées ($X_p$, $Y_p$, $Z_p$) exprimées dans le repère $R_0$ :

$$\mathbf{L} = \mathbf{F}(^0\mathbf{P}_p) \qquad (2)$$

avec $^0\mathbf{P}_p = \begin{bmatrix} X_P & Y_P & Z_P \end{bmatrix}^T$

Le modèle géométrique inverse est de la forme suivante :

$$\begin{bmatrix} q_{11} \\ q_{12} \\ q_{13} \end{bmatrix} = \begin{bmatrix} Z_P - \cos(q_{31})\cos(q_{21})D_{41} - D_{61} \\ X_P - X_{A2} - \cos(q_{32})\cos(q_{22})D_{42} - D_{62} \\ Y_P - Y_{A3} - \cos(q_{33})\cos(q_{23})D_{43} - D_{63} \end{bmatrix} \qquad (3)$$

avec :

$q_{31} = \sin^{-1}\left(\frac{-Y_P}{D_{41}}\right)$ et $q_{21} = -\left(\sin^{-1}\left(\frac{-X_P}{\cos(q_{31})D_{41}}\right) + \frac{\pi}{2}\right)$ ;

$q_{32} = \sin^{-1}\left(\frac{-Z_P + Z_{A2}}{D_{42}}\right)$ et $q_{22} = -\left(\sin^{-1}\left(\frac{-Y_P + Y_{A2}}{\cos(q_{32})D_{42}}\right) + \frac{\pi}{2}\right)$ ;

$q_{33} = \sin^{-1}\left(\frac{-X_P + X_{A3}}{D_{43}}\right)$ et $q_{23} = -\left(\sin^{-1}\left(\frac{-Z_P + Z_{A3}}{\cos(q_{33})D_{43}}\right) + \frac{\pi}{2}\right)$ ;

ii) Le modèle cinématique inverse du robot donne la vitesse des variables prismatiques motorisées ($\dot{q}_{11}, \dot{q}_{12}, \dot{q}_{13}$) en fonction de la vitesse de la plate-forme :

$$\dot{\mathbf{L}} = {}^0\mathbf{J}_p^{-1}\, {}^0\mathbf{V}_p \qquad (4)$$

où $^0\mathbf{J}_p^{-1}$ est la matrice Jacobienne inverse du robot :

$$^0\mathbf{J}_p^{-1} = \begin{bmatrix} -\dfrac{1}{T_{21}} & \dfrac{T_{31}}{S_{21}} & 1 \\ 1 & -\dfrac{1}{T_{22}} & \dfrac{T_{32}}{S_{22}} \\ \dfrac{T_{33}}{S_{23}} & 1 & -\dfrac{1}{T_{23}} \end{bmatrix} \qquad (5)$$

avec : $T_{2i} = \tan(q_{2i})$, $S_{3i} = \sin(q_{3i})$ et $T_{3i} = \tan(q_{3i})$.

iii) Le modèle cinématique inverse de la chaîne i donne la vitesse des articulations de la chaîne i ($\dot{q}_{1i}, \dot{q}_{2i}, \dot{q}_{3i}$) en fonction de la vitesse du point P :

$$\dot{\mathbf{q}}_i = {}^0\mathbf{J}_i^{-1}\, {}^0\mathbf{V}_p \qquad (6)$$

où $^0\mathbf{J}_i^{-1}$ est la matrice Jacobienne inverse de la chaîne i.

$$^0\mathbf{J}_1 = \begin{bmatrix} 0 & -D_{41}C_{31}S_{21} & -D_{41}S_{31}C_{21} \\ 0 & 0 & -D_{41}C_{31} \\ 1 & -D_{41}C_{31}C_{21} & D_{41}S_{31}S_{21} \end{bmatrix}, \quad ^0\mathbf{J}_2 = \begin{bmatrix} 1 & -D_{42}C_{32}C_{22} & D_{42}S_{32}S_{22} \\ 0 & -D_{42}C_{32}S_{22} & -D_{42}S_{32}C_{22} \\ 0 & 0 & -D_{42}C_{32} \end{bmatrix}$$

$$^0\mathbf{J}_3 = \begin{bmatrix} 0 & 0 & -D_{43}C_{33} \\ 1 & -D_{43}C_{33}C_{23} & D_{43}S_{33}S_{23} \\ 0 & -D_{43}C_{33}S_{23} & -D_{43}S_{33}C_{23} \end{bmatrix} \qquad (7)$$

$$^0\mathbf{J}_1^{-1} = \begin{bmatrix} -\dfrac{1}{T_{21}} & \dfrac{T_{31}}{S_{21}} & 1 \\ -\dfrac{1}{D_{41}C_{31}S_{21}} & \dfrac{T_{31}}{D_{41}C_{31}T_{21}} & 0 \\ 0 & -\dfrac{1}{D_{41}C_{31}} & 0 \end{bmatrix}, \quad ^0\mathbf{J}_2^{-1} = \begin{bmatrix} 1 & -\dfrac{1}{T_{22}} & \dfrac{T_{32}}{S_{22}} \\ 0 & -\dfrac{1}{D_{42}C_{32}S_{22}} & \dfrac{T_{31}}{D_{42}C_{32}T_{22}} \\ 0 & 0 & -\dfrac{1}{D_{42}C_{32}} \end{bmatrix}$$

$$^0\mathbf{J}_3^{-1} = \begin{bmatrix} \dfrac{T_{33}}{S_{23}} & 1 & -\dfrac{1}{T_{23}} \\ \dfrac{T_{33}}{D_{43}C_{33}T_{23}} & 0 & -\dfrac{1}{D_{43}C_{33}S_{23}} \\ -\dfrac{1}{D_{43}C_{33}} & 0 & 0 \end{bmatrix} \qquad (8)$$

iv) Le modèle cinématique inverse du second ordre de la chaîne i donne l'accélération des articulations de la chaîne i ($\ddot{q}_{1i}, \ddot{q}_{2i}, \ddot{q}_{3i}$) en fonction de l'accélération du point P :

$$\ddot{\mathbf{q}}_i = {}^0\mathbf{J}_i^{-1}\left({}^0\dot{\mathbf{V}}_p - {}^0\dot{\mathbf{J}}_i\dot{\mathbf{q}}_i\right) \qquad (9)$$

## IV. MODÈLE DYNAMIQUE INVERSE

Le modèle dynamique inverse représente la relation entre les forces des articulations motorisées et les positions, vitesses, accélérations cartésiennes de la plate-forme. Il s'écrit $\Gamma = \mathbf{f}\left({}^0\mathbf{P}_P, {}^0\mathbf{V}_p, {}^0\dot{\mathbf{V}}_p\right)$.

En utilisant les modèles géométrique et cinématique inverse du robot et des chaînes on peut exprimer la position, la vitesse et l'accélération des articulations de chaque chaîne ($\mathbf{q}_i, \dot{\mathbf{q}}_i, \ddot{\mathbf{q}}_i$) en fonction du mouvement de la plate-forme $\left({}^0\mathbf{P}_P, {}^0\mathbf{V}_p, {}^0\dot{\mathbf{V}}_p\right)$.

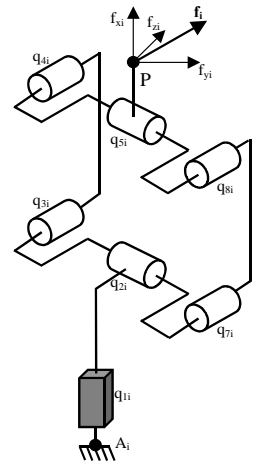

Fig. 5. : Force $f_i$

En représentant les réactions de la plate-forme sur la chaîne i par $\mathbf{f}_i$ (figure 5), les inconnues du modèle dynamique inverse sont les 9 composantes des forces $\mathbf{f}_i = \begin{bmatrix} f_{xi} & f_{yi} & f_{zi} \end{bmatrix}^T$ et les 3 forces des articulations indépendantes $\Gamma_{1i}$ (i = 1, 2, 3). Le modèle dynamique inverse de chaque chaîne est composée de 3 équations indépendantes (cf : annexe) ce qui donne un total de 9 équations. Les équations de Newton-Euler de la plate-forme donnent 3 équations supplémentaires. Le système est donc composé de 12 inconnues et de 12 équations, il est donc envisageable d'être résolu. Dans la suite, ce système d'équations sera résolu de façon séquentielle.

### A. Calcul de $^0\mathbf{f_i}$ en utilisant le modèle dynamique de la chaîne i

La forme générale du modèle dynamique de la chaîne i, s'écrit :

$$\mathbf{\Gamma_i} = \mathbf{H_i}(\mathbf{q_i},\dot{\mathbf{q}}_i,\ddot{\mathbf{q}}_i) + {}^0\mathbf{J}_i^T\,{}^0\mathbf{f_i} \tag{10}$$

où :

$$\mathbf{H_i}(\mathbf{q_i},\dot{\mathbf{q}}_i,\ddot{\mathbf{q}}_i) = \mathbf{A_i}\ddot{\mathbf{q}}_i + \mathbf{h_i}(\mathbf{q_i},\dot{\mathbf{q}}_i) \tag{11}$$

$\mathbf{A_i}$ est la matrice d'inertie (3×3) de la chaîne i et $\mathbf{h_i}$ est la matrice (3×1) contenant les forces de Coriolis, centrifuge et gravité. $\mathbf{\Gamma_i}$ est composé des couples/forces de la chaîne i, où $\Gamma_{1i}$ et $\Gamma_{2i}$ sont nuls :

$$\mathbf{\Gamma_i} = \begin{bmatrix} \Gamma_{1i} & \Gamma_{2i} & \Gamma_{3i} \end{bmatrix}^T = \begin{bmatrix} \Gamma_{1i} & 0 & 0 \end{bmatrix}^T \tag{12}$$

En utilisant l'équation (10), les forces de réaction $^0\mathbf{f_i}$ peuvent s'écrire :

$${}^0\mathbf{f_i} = {}^0\mathbf{J}_i^{-T}\left(\mathbf{\Gamma_i} - \mathbf{H_i}(\mathbf{q_i},\dot{\mathbf{q}}_i,\ddot{\mathbf{q}}_i)\right) \tag{13}$$

$${}^0\mathbf{f_i} = -{}^0\mathbf{J}_i^{-T}\mathbf{H_i}(\mathbf{q_i},\dot{\mathbf{q}}_i,\ddot{\mathbf{q}}_i) + {}^0\mathbf{J}_i^{-T}\mathbf{\Gamma_i} \tag{14}$$

En notant $^0\mathbf{J}_i^{-T}\mathbf{H_i}(\mathbf{q_i},\dot{\mathbf{q}}_i,\ddot{\mathbf{q}}_i)$ par $\mathbf{H_{xi}}(\mathbf{q_i},\dot{\mathbf{q}}_i,\ddot{\mathbf{q}}_i)$, alors :

$${}^0\mathbf{f_i} = -\mathbf{H_{xi}}(\mathbf{q_i},\dot{\mathbf{q}}_i,\ddot{\mathbf{q}}_i) + {}^0\mathbf{J}_i^{-T}\mathbf{\Gamma_i} \tag{15}$$

$\mathbf{H_{xi}}(\mathbf{q_i},\dot{\mathbf{q}}_i,\ddot{\mathbf{q}}_i)$ exprime le vecteur $\mathbf{H_i}(\mathbf{q_i},\dot{\mathbf{q}}_i,\ddot{\mathbf{q}}_i)$ dans l'espace cartésien de position au point $P_i$ [18,19].

En utilisant l'équation (8), on obtient pour la chaîne 1 :

$${}^0\mathbf{f_1} = -\mathbf{H_{x1}}(\mathbf{q_1},\dot{\mathbf{q}}_1,\ddot{\mathbf{q}}_1) + \begin{bmatrix} -\dfrac{1}{T_{21}} \\ \dfrac{T_{31}}{S_{21}} \\ 1 \end{bmatrix}\Gamma_{11} \tag{16}$$

On peut observer que le coefficient de $\Gamma_{11}$ dans l'équation (16) correspond à la première colonne de la matrice Jacobienne inverse transposée du robot (équation (5)). Il en est de même pour les chaînes 2 et 3, qui donne des matrices respectivement fonction de $\Gamma_{12}$ et $\Gamma_{13}$, qui correspondent respectivement à la deuxième et la troisième colonne de la matrice Jacobienne inverse transposée du robot :

$${}^0\mathbf{f_i} = -\mathbf{H_{xi}}(\mathbf{q_i},\dot{\mathbf{q}}_i,\ddot{\mathbf{q}}_i) + {}^0\mathbf{J}_{p\,[:,i]}^{-T}\mathbf{\Gamma_i} \tag{17}$$

Où ${}^0\mathbf{J}_{p\,[:,i]}^{-T}$ représente la i$^{ème}$ colonne de la matrice Jacobienne inverse transpose du robot.

### B. Dynamique de la plate-forme

Les équations de Newton-Euler appliquées à l'origine de la plate-forme s'écrivent (pas de rotation) :

$${}^0\mathbf{F_P} = {}^0\dot{\mathbf{V}}_P\,M_p - M_p\,{}^0\mathbf{g} \tag{18}$$

Avec :
$^0\mathbf{g}$ accélération de la pesanteur exprimée dans le repère $R_0$
$M_p$ masse de la plate-forme
$^0\mathbf{F_P}$ Force totale extérieure appliquée sur la plate-forme au point P. Puisque $^0\dot{\mathbf{V}}_P$ est connu, alors $^0\mathbf{F_p}$ peut-être calculé à partir de l'équation (18).

Les forces appliquées à la plate-forme dues aux forces de réaction de chaque chaîne (figure 6) sont calculées en utilisant l'équation suivante :

$${}^0\mathbf{F_p} = \sum_{i=1}^{3}{}^0\mathbf{f_i} \tag{19}$$

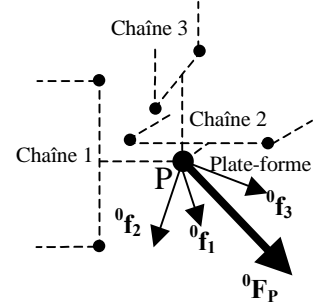

Fig. 6. : Forces extérieures appliquées au point P

En substituant l'équation (17) dans l'équation (19), on obtient :

$${}^0\mathbf{F_p} = \sum_{i=1}^{3}\left[\left(-\mathbf{H_{xi}}(\mathbf{q_i},\dot{\mathbf{q}}_i,\ddot{\mathbf{q}}_i) + {}^0\mathbf{J}_{p\,[:,i]}^{-T}\Gamma_{1i}\right)\right] \tag{20}$$

où :

$${}^0\mathbf{J}_p^{-T}\,\mathbf{\Gamma_{robot}} = {}^0\mathbf{F_p} + \sum_{i=1}^{3}\left[\mathbf{H_{xi}}(\mathbf{q_i},\dot{\mathbf{q}}_i,\ddot{\mathbf{q}}_i)\right] \tag{21}$$

On peut en déduire :

$$\mathbf{\Gamma_{robot}} = {}^0\mathbf{J}_p^{T}\mathbf{H_{robot}} \tag{22}$$

avec :

$$\mathbf{\Gamma_{robot}} = \begin{bmatrix} \Gamma_{11} & \Gamma_{12} & \Gamma_{13} \end{bmatrix}^T$$

$$\mathbf{H_{robot}} = {}^0\mathbf{F_p} + \sum_{i=1}^{3}\left[\mathbf{H_{xi}}(\mathbf{q_i},\dot{\mathbf{q}}_i,\ddot{\mathbf{q}}_i)\right] \tag{23}$$

L'équation (22) représente le modèle dynamique inverse. Pour le calculer, en plus du calcul des matrices Jacobiennes, nous devons déterminer $\mathbf{H_i}(\mathbf{q_i},\dot{\mathbf{q}}_i,\ddot{\mathbf{q}}_i)$ qui représente le modèle dynamique inverse de la chaîne i. Différentes méthodes peuvent être utilisées pour calculer ce vecteur numériquement ou symboliquement [20,21,22]. Afin de réduire les coûts du calcul, la méthode de Newton-Euler s'appuyant sur un calcul symbolique itératif et utilisant les paramètres inertiels de base du robot peut-être utilisée [17,23].

On peut remarquer que le modèle dynamique inverse, équation (22), ne possède pas de singularités dans l'espace accessible du robot.

### C. Interprétation physique

i) On peut observer sur l'équation (23), que l'effet des chaînes cinématiques sur la plate-forme équivaut à appliquer une force égale à $\mathbf{H_{xi}}(\mathbf{q_i},\dot{\mathbf{q}}_i,\ddot{\mathbf{q}}_i)$ au point P. La force totale correspondante est égale à $\sum\mathbf{H_{xi}}(\mathbf{q_i},\dot{\mathbf{q}}_i,\ddot{\mathbf{q}}_i)$.

ii) Le membre à gauche de l'équation (21) représente les forces des actionneurs exprimées dans l'espace cartésien de la plate-forme.

## V. MODÈLE DYNAMIQUE DIRECT

Le modèle dynamique direct exprime les accélérations cartésiennes de la plate-forme en fonction des positions et vitesses de la plate-forme et les forces des articulations motorisées. Il s'écrit : $^0\dot{\mathbf{V}}_P = \mathbf{f}\left({}^0\dot{\mathbf{V}}_P,\,{}^0\mathbf{V}_P,\,\mathbf{\Gamma}\right)$.

Les inconnues pour ce problème sont les 9 composantes des forces de réaction au point P sur chaque chaîne $\mathbf{f_i} = [f_{xi} \quad f_{yi} \quad f_{zi}]^T$ et les 3 composantes de l'accélération linéaire cartésienne de la plate-forme $^0\mathbf{V_p}$. Comme pour le modèle dynamique inverse, on a 9 équations obtenues à partir des modèles dynamiques inverses des chaînes et 3 équations provenant de la plate-forme. Le système est donc composé également de 12 inconnues et de 12 équations. Ce système d'équations sera résolu de façon séquentielle.

### A. Calcul de $^0\mathbf{f_i}$ en utilisant le modèle dynamique de la chaîne i.

Les forces de réaction de la chaîne, relation (14), doivent être exprimées en fonction des accélérations linéaires cartésiennes de la plate-forme. En utilisant les relations (11) et (17) de façon à séparer les forces d'inerties des forces de Coriolis, centrifuges et gravités, on obtient :

$$^0\mathbf{f_i} = -{}^0\mathbf{J_i^{-T}}\mathbf{A_i}\ddot{\mathbf{q}}_i - {}^0\mathbf{J_i^{-T}}\mathbf{h_i}(\mathbf{q_i},\dot{\mathbf{q}}_i) + {}^0\mathbf{J_{p[:,i]}^{-T}}\Gamma_i \quad (24)$$

En substituant l'équation (9) dans (24), on a :
$$^0\mathbf{f_i} = -\mathbf{A_{xi}}\,{}^0\dot{\mathbf{V}}_p + \mathbf{A_{xi}}\,{}^0\dot{\mathbf{J}}_i\dot{\mathbf{q}}_i - \mathbf{h_{xi}}(\mathbf{q_i},\dot{\mathbf{q}}_i) + {}^0\mathbf{J_{p[:,i]}^{-T}}\Gamma_i \quad (25)$$

Où $\mathbf{A_{xi}} = {}^0\mathbf{J_i^{-T}}\mathbf{A_i}\,{}^0\mathbf{J_i^{-1}}$ est la matrice (3×3) d'inertie cartésienne de la chaîne i exprimée au point $P_i$, et $\mathbf{h_{xi}} = {}^0\mathbf{J_i^{-T}}\mathbf{h_i}$ est la matrice (3×1) de Coriolis, centrifuge et gravité exprimée dans l'espace cartésien au point $P_i$ [18,19].

En substituant l'équation (25) dans les équations de Newton-euler de la plate-forme, relation (18), on obtient :

$$^0\mathbf{F_p} = -\sum_{i=1}^{3}[\mathbf{A_{xi}}]\,{}^0\dot{\mathbf{V}}_p + \sum_{i=1}^{3}[\mathbf{A_{xi}}\,{}^0\dot{\mathbf{J}}_i\dot{\mathbf{q}}_i - \mathbf{h_{xi}}(\mathbf{q_i},\dot{\mathbf{q}}_i)] + {}^0\mathbf{J_P^{-T}}\Gamma_{robot} \quad (26)$$

Ce qui s'écrit :
$$\left(\sum_{i=1}^{3}[\mathbf{A_{xi}}] + \mathbf{I_3}M_P\right){}^0\dot{\mathbf{V}}_p + \sum_{i=1}^{3}[\mathbf{h_{xi}}(\mathbf{q_i},\dot{\mathbf{q}}_i) - \mathbf{A_{xi}}\,{}^0\dot{\mathbf{J}}_i\dot{\mathbf{q}}_i] - M_P\mathbf{g} = {}^0\mathbf{J_P^{-T}}\Gamma_{robot} \quad (27)$$

Les accélérations cartésiennes de la plate-forme peuvent être calculées à partir de la relation suivante suivante :
$$^0\dot{\mathbf{V}}_{pi} = \mathbf{A_{robot}^{-1}}\left({}^0\mathbf{J_P^{-T}}\Gamma_{robot} - \mathbf{h_{robot}}\right) \quad (28)$$

Avec :
$$\Gamma_{robot} = [\Gamma_{11} \quad \Gamma_{12} \quad \Gamma_{13}]^T$$
$$\mathbf{A_{robot}} = \sum_{i=1}^{3}[\mathbf{A_{xi}}] + \mathbf{I_3}M_p \quad (29)$$
$$\mathbf{h_{robot}} = \sum_{i=1}^{3}[\mathbf{h_{xi}}(\mathbf{q_i},\dot{\mathbf{q}}_i) - \mathbf{A_{xi}}\,{}^0\dot{\mathbf{J}}_i\dot{\mathbf{q}}_i] - M_p\mathbf{g} \quad (30)$$

Les accélérations articulaires $\ddot{q}_{1i}, \ddot{q}_{2i}, \ddot{q}_{3i}$ (pour i = 1 à 3) des trois chaînes cinématiques peuvent être obtenues en utilisant le modèle cinématique inverse du second ordre (relation (9)).

La matrice (3×3) $\mathbf{A_{robot}}$ correspond à la somme des matrices d'inerties spatiales des chaînes plus la matrice (3×3) diagonale dont les éléments sont égaux à $M_P$.

L'équation (28) représente le modèle dynamique direct du robot. Pour l'obtenir, on calcule les termes $\mathbf{A_i}$ et $\mathbf{h_i}$ de chaque chaîne :
- $\mathbf{h_i}$ peut être obtenu en utilisant la formulation de Newton-Euler dans le modèle dynamique inverse de chaque chaîne en considérant les accélérations $\ddot{\mathbf{q}}_i$ nulles : $\mathbf{h_i} = \mathbf{H_i}(\mathbf{q_i},\dot{\mathbf{q}}_i,\ddot{\mathbf{q}}_i = 0)$ [24,25].

- $\mathbf{A_i}$ peut être obtenu en utilisant la formulation de Newton-Euler [24,25] ou Lagrange [17].

Pour réduire les coûts de calcul, on peut utiliser une méthode de calcul symbolique itératif et les paramètres inertiels de base du robot pour $\mathbf{A_i}$ et $\mathbf{h_i}$ [25].

La relation (28) peut être aussi utiliser pour le calcul du modèle dynamique inverse, mais la relation (22) est plus efficace du point de vue coût de calcul.

### B. Interprétation physique

i) La contribution de chaque chaîne sur la matrice d'inertie spatiale du robot $\mathbf{A_{robot}}$ est due à la matrice (3×3) d'inertie cartésienne de la chaîne exprimée au point P.

ii) La matrice (3×1) $\mathbf{h_{xi}}$ représente les forces de Coriolis, centrifuges et gravité exprimées dans l'espace cartésien au point P. Les forces totales correspondantes sont données par $\sum(\mathbf{h_{xi}}(\mathbf{q_i},\dot{\mathbf{q}}_i) - \mathbf{A_{xi}}\,{}^0\dot{\mathbf{J}}_i\dot{\mathbf{q}}_i)$.

## VI. CONCLUSION

Dans cet article, nous présentons une forme intéressante des modèles dynamiques inverse et direct du robot Orthoglide. En effet, ces modèles prennent en compte toute la dynamique du robot sans simplification et permettent de déduire une interprétation physique du modèle dynamique du robot parallèle. Cette approche pourra être utilisée sur d'autres structures parallèles.

Ces modèles sont obtenus en termes des éléments du modèle dynamique cartésien des chaînes cinématiques du robot perçues aux points de connexions des chaînes avec la plate-forme. Par conséquent, on peut appliquer les techniques développées pour les robots séries, aux calculs de ces modèles.

ANNEXE

*Calcul du modèle dynamique inverse de la chaîne i*

Chaque chaîne du robot est composée d'une boucle planaire de type parallélogramme. On calcule le modèle dynamique de la structure arborescente équivalente en coupant l'articulation passive ($q_{8i}$) du parallélogramme (figure 3) :

$$\Gamma_{ar_i} = H_{ar_i}(q_i, \dot{q}_i, \ddot{q}_i) \quad (31)$$

avec $\Gamma_{ar_i} = \begin{bmatrix} \Gamma_{1i} & \Gamma_{2i} & \Gamma_{3i} & \Gamma_{4i} & \Gamma_{5i} & \Gamma_{7i} \end{bmatrix}^T$

On calcule ensuite les équations de contrainte de fermeture de boucle pour exprimer les variables des articulations passives en fonction des variables des articulations actives. La chaîne i étant isolée de la plate-forme, on considère dans un premier temps que les articulations $q_{1i}$, $q_{2i}$ et $q_{3i}$ sont actives. Dans le modèle complet du robot, on rappelle que seule $q_{1i}$ est motorisée et que l'on considère les couples $\Gamma_{2i}$ et $\Gamma_{3i}$ de valeurs nulles.

Soit $q_{a_i}$ le vecteur contenant les articulations actives et $q_{p_i}$ le vecteur contenant les articulations passives :

$$q_{a_i} = \begin{bmatrix} q_{1i} & q_{2i} & q_{3i} \end{bmatrix}^T \quad (32)$$

$$q_{p_i} = \begin{bmatrix} q_{4i} & q_{5i} & q_{7i} \end{bmatrix}^T \quad (33)$$

Les équations de fermeture de boucle sont les suivantes :

$$\left. \begin{aligned} q_{4i} &= -q_{3i} \\ q_{5i} &= -q_{2i} - \frac{\pi}{2} \\ q_{7i} &= q_{3i} \\ q_{8i} &= -q_{3i} \end{aligned} \right\} \quad (34)$$

Le modèle dynamique de la boucle fermée est obtenu à partir de $\Gamma_{ar_i}$ et des équations de contraintes en utilisant la relation suivante [17] :

$$\Gamma_i = \begin{bmatrix} \Gamma_{1i} & \Gamma_{2i} & \Gamma_{3i} \end{bmatrix}^T = G_i^T \Gamma_{ar_i} \quad (35)$$

$$\Gamma_i = G_i^T H_{ar_i}(q_i, \dot{q}_i, \ddot{q}_i) = H_i(q_i, \dot{q}_i, \ddot{q}_i) \quad (36)$$

avec : $G_i = \dfrac{\partial q_{ar_i}}{\partial q_{a_i}}$ et $q_{ar_i} = \begin{bmatrix} q_{a_i} & q_{p_i} \end{bmatrix}$

En utilisant les équations (34), on obtient :

$$G^T = \begin{bmatrix} 1 & 0 & 0 & 0 & 0 & 0 \\ 0 & 1 & 0 & 0 & -1 & 0 \\ 0 & 0 & 1 & -1 & 0 & 1 \end{bmatrix} \quad (37)$$

$H_i(q_i, \dot{q}_i, \ddot{q}_i)$ est le modèle dynamique de la chaîne i. Lorsque l'on prend en compte la force de réaction $f_i$ sur le point terminal de chaque chaîne, la forme générale du modèle dynamique de la chaîne i, devient :

$$\Gamma_i = H_i(q_i, \dot{q}_i, \ddot{q}_i) + {}^0J_i^T\, {}^0f_i \quad (38)$$